\documentclass[twocolumn,journal]{IEEEtran}
\usepackage[T1]{fontenc}
\usepackage{color}
\usepackage{mathrsfs}
\usepackage{amsmath}
\usepackage{amsthm}
\usepackage{amssymb}
\usepackage{graphicx}
\usepackage[colorlinks=false]{hyperref}
\hypersetup{pdftitle={Your Title},pdfauthor={Your Name},pdfpagelayout=OneColumn,pdfnewwindow=true, pdfstartview=XYZ, plainpages=false}
\usepackage{breakurl}

\makeatletter

\providecommand{\tabularnewline}{\\}

 \let\oldforeign@language\foreign@language
 \DeclareRobustCommand{\foreign@language}[1]{%
   \lowercase{\oldforeign@language{#1}}}
\theoremstyle{plain}
\newtheorem{thm}{\protect\theoremname}
\theoremstyle{plain}
\newtheorem{lem}[thm]{\protect\lemmaname}

\usepackage[caption=false,font=footnotesize]{subfig}

\makeatother

\providecommand{\lemmaname}{Lemma}
\providecommand{\theoremname}{Theorem}

\begin{document}

\title{Estimation of respiratory pattern from video using selective ensemble
aggregation}

\author{Prathosh AP, \IEEEmembership{Member,~IEEE}, Pragathi Praveena, \IEEEmembership{Member,~IEEE},
Lalit K Mestha, \IEEEmembership{Fellow,~IEEE}, Sanjay Bharadwaj\thanks{Prathosh AP and Pragathi Praveena are with Xerox Research Center India
e-mail:\{prathosh.ap,pragathi.praveena\}@Xerox.com and prathoshap@gmail.com.
Lalit K Meshta is with GE global research, Niskayuna, USA. e-mail:
lalit.mestha@ge.com. Sanjay Bharadwaj is with Skanray Technologies,
Mysore, India. e-mail: Sanjay.bharadwaj@skanray.com. }}

\IEEEspecialpapernotice{}

\IEEEaftertitletext{}

\markboth{IEEE Transactions on Signal Processing}{Your Name \MakeLowercase{\emph{et al.}}: Your Title}

\IEEEpubid{}
\maketitle
\begin{abstract}
Non-contact estimation of respiratory pattern (RP) and respiration
rate (RR) has multiple applications. Existing methods for RP and RR
measurement fall into one of the three categories - (i) estimation
through nasal air flow measurement, (ii) estimation from video-based
remote photoplethysmography, and (iii) estimation by measurement of
motion induced by respiration using motion detectors. These methods,
however, require specialized sensors, are computationally expensive
and/or critically depend on selection of a region of interest (ROI)
for processing. In this paper a general framework is described for
estimating a periodic signal driving noisy LTI channels connected
in parallel with unknown dynamics. The method is then applied to derive
a computationally inexpensive method for estimating RP using 2D cameras
that does not critically depend on ROI. Specifically, RP is estimated
by imaging the changes in the reflected light caused by respiration-induced
motion. Each spatial location in the field of view of the camera is
modeled as a noise-corrupted linear time-invariant (LTI) measurement
channel with unknown system dynamics, driven by a single generating
respiratory signal. Estimation of RP is cast as a blind deconvolution
problem and is solved through a method comprising subspace projection
and statistical aggregation. Experiments are carried out on 31 healthy
human subjects by generating multiple RPs and comparing the proposed
estimates with simultaneously acquired ground truth from an impedance
pneumography device. The proposed estimator agrees well with the ground
truth device in terms of correlation measures, despite variability
in clothing pattern, angle of view and ROI. 
\end{abstract}


\IEEEpeerreviewmaketitle{}

\section{Introduction}

\subsection{Background }

Respiration is a fundamental physiological activity \cite{joanna2009vital}
and is associated with several muscular, neural and chemical processes
within the body of living organisms. Given the fact that respiratory
diseases such as chronic obstructive pulmonary disease, asthma, tuberculosis,
sleep apnea and respiratory tract infections account for about 18\%
of human deaths worldwide \cite{WHO_report}, assessment of multiple
respiratory parameters is of major importance for diagnosis and monitoring.
Accordingly, respiratory parameters such as respiration rate (RR),
respiration pattern (RP) and respiratory flow-volume are routinely
measured in clinical and primary healthcare settings. RR refers to
the number of inhalation-exhalation cycles (breaths) observed per
unit time, usually quantified as breaths per minute (BPM). RP refers
to a temporal waveform signifying multiple phases of the respiratory
function such as intervals and peaks of inhalation and exhalation,
relative amplitudes of different breath cycles and cycle frequency
(instantaneous RR). Respiratory flow-volume measures the amount of
air that is inhaled/exhaled in every breath. 

A simple application of RR measurement is in assessing whether a human
is breathing or not. Apart from this, deviation from the permissible
RR range (usually 6-35 breaths per minute in healthy adults) signifies
pulmonary and cardiac abnormalities \cite{yasuma2004respiratory}.
For example, abnormally high RR is symptomatic of diseases like pneumonia
in children. Further, estimation of RP has several applications such
as detection of sleep apnea, gating signal generation \cite{kubo1996respiration}
for medical imaging and psychological state assessment. Sleep apnea
is characterized by disrupted breathing patterns (cessation, shallowing,
flow-blockage) during sleep which can be detected if a reliable estimate
of RP is available. RP is used in respiration gated image acquisition,
where radiographic images of human anatomical regions are captured
synchronously with certain significant points of the RP (for instance,
an image is acquired at every inspiration peak) to facilitate accurate
image registration and minimize exposure to harmful X-rays. A similar
technique is used in therapeutic energy delivery methods such as lithotripsy
where shock waves are administered to posterior lower back region
at certain temporal triggers of the RP. Further, different RPs indicate
different sympathetic and parasympathetic responses leading to potential
analysis of human emotions such as anger and stress. 

Given their aforementioned significance, accurate estimation of RP
and RR has been considered an important problem in the biomedical
engineering community for decades. Several accurate and robust techniques
such as spirometry \cite{miller2005standardisation}, impedance pneumography
\cite{geddes1962impedance} and plethysmography \cite{nakajima1996monitoring}
can measure RR and some can also estimate RP. However, they employ
contact-based leads, straps and probes which may not be optimal for
use in situations such as neonatal ICU, home health monitoring and
gated image acquisition. This is due to several reasons such as sensitive
skin, discomfort or irritation and interference of leads with the
radiographic images acquired. Owing to such needs, a recent trend
in non-contact respiratory monitoring has emerged. In the following
section, a brief review of non-contact methods for respiratory monitoring
is provided. 

\subsection{Prior work }

Existing methods for non-contact RP estimation fall into one of the
three categories - (i) estimation through indirect nasal air flow
measurement, (ii) estimation by imaging volumetric changes in blood
using remote photoplethysmography, and (iii) estimation by measurement
of motion induced due to respiration. In the first category, the idea
is to indirectly measure the amount of air inhaled and exhaled during
each cycle using different modalities. One technique is phonospirometry
\cite{macklem2001phonospirometry,que2002phonospirometry}, where the
respiratory parameters are estimated from measurements of tracheal
breath sounds captured using acoustic microphones placed near the
trachea. Based on the observation that the air exhaled has a higher
temperature than the typical background of indoor environments, there
are attempts to measure breathing function using highly sensitive
infrared imaging \cite{murthy2006noncontact,al2015tracing}. These
two methods demand sensitive microphones and thermal imaging systems
as additional hardware. Also, it has been noted that subtle breathing
is hard to measure using phonospirometry.

The second category of algorithms are based on the observation that
respiration information rides over the photoplethesmogram (PPG) signal
as an amplitude modulation component. A gamut of recent works concentrate
on camera-based PPG estimation \cite{verkruysse2008remote,chon2009estimation,poh2011advancements}.
The basic idea in all these is to capture the subtle changes in skin
color occurring from pulsatile changes in arterial blood volume in
human body tissues. It is well recognized that these methods (often
called remote PPG or rPPG) are highly sensitive to subject motion,
skin color and ambient light. A lot of effort has been put in improving
the robustness of rPPG against these artifacts and significant progress
has been made using several signal processing and statistical modeling
techniques including blind source separation \cite{poh2010non}, alternative
reflectance models, spatial pruning \cite{wang2015exploiting}, temporal
filtering \cite{nakajima1996monitoring} and autoregressive modeling
\cite{FlemingTarassenko2007}. These methods albeit mature can only
provide an estimate of RR but cannot estimate RP. Further, in some
cases, they require a careful selection of a region of interest (often
facial region) for processing. 

The third category of methods rely on measuring the motion induced
in different body parts due to respiration. One proposed method \cite{min2010noncontact}
is to use an ultrasonic proximity sensor (typically mounted on a stand
placed in front of the subject) to measure the chest-wall motion induced
by respiration. Techniques based on (a) laser diodes measuring the
distance between the chest wall and the sensor \cite{kondo1997laser}
and (b) Doppler radar system measuring the Doppler shift in the transmitted
waves induced by respiratory chest wall motion \cite{mabrouk2014model,gu2015assessment}
are also proposed. These methods demand dedicated sensors and in some
cases have been reported to depend on the texture of the cloth on
the subject. Some methods \cite{yu2012noncontact,benetazzo2014respiratory}
employ depth sensing cameras (such as Kinect) \cite{bernal2014non}
to directly measure the variations in the distance between a fixed
surface (such as wall) and the chest-wall. There have been few attempts
in estimating the RP using consumer grade 2D cameras: an attempt has
been made by Shao \textit{et al.} \cite{shao2014noncontact}, where
the upward and downward motion in the shoulders due to the respiration
is measured using differential signal processing, which is highly
sensitive to the selection of region of interest (ROI) comprising
the shoulder region. Very recently, use of Haar-like features derived
from optical flow vectors computed on the chest region is proposed
to estimate RR \cite{lin2016image}. Janssen et.al \cite{janssen2015video}
proposes an automatic ROI selection method for RP estimation based
on the observation that the respiration-induced chest-wall motion
is uncorrelated from the remaining sources. The idea is to extract
the dense optical flow vectors in the entire scene followed by a robust
feature representation exploiting the intrinsic properties of respiration.
These features are then factorized to get the respiration signal.
One of our recent techniques also falls into this category \cite{avishek}.
These methods are shown to be accurate and robust, however, they require
computation of optical flow field for multiple frames which is known
to be computationally expensive. In this paper, we propose a method
to estimate the respiration pattern and rate using a consumer grade
2D camera. The method is computationally inexpensive and does not
critically depend on the texture of the cloth, angle of view of the
camera and selection of ROI. 

\subsection{Premise and objectives}

Suppose a consumer grade camera is placed in front of a steady human
subject such that its field-of-view comprises the abdominal-thoracic
region of the subject. Assume that the relative position of the camera
with respect to the subject does not change and also that the luminance
of the background lighting is fairly constant\footnote{These are reasonable assumptions in many uses cases such as respiration
gated image acquisition, non-contact monitoring and RR estimation,
where the cameras are held fairly stable in a bright environment.
Cases where there are dominant relative motion, fluctuation in the
lighting are separate problems by themselves and hence beyond the
scope of the present paper. }. Under such conditions, if a subject's abdominal-thoracic region
is imaged using a video during breathing, the changes in each pixel
value measured will be a function of the motion induced by respiration
and the surface reflectance characteristics of the region imaged.
Since each pixel response is distinct, the core problem of RP estimation
can be posed as the following:\textit{ How to process individual pixel
responses to obtain the respiratory pattern? }

This problem is solved by modeling every pixel as the output of a
linear time invariant (LTI) channel of unknown system response driven
by a hypothetical generating respiration signal that is to be estimated.
The problem of estimation of RP is cast as the following estimation
problem: Estimate the input signal, given the outputs of several independent
noise-corrupted LTI channels with unknown system responses that are
driven by the same generating input signal. This is referred to as
the blind deconvolution problem of the single-input multiple-output
(SIMO) systems in the signal processing community which is often solved
through an assumed parametric form for the input signal and/or the
system responses followed by error minimization techniques defined
on different cost functions \cite{zhang2004multichannel,makhoul1975linear,kundur1998novel,moulines1995subspace,tong1994blind}.
However, in this paper, we propose a solution for blind deconvolution
of periodic signals with a certain class of system characteristics
where we neither assume any form for the transfer functions of the
individual systems nor rely on error minimization.

\section{Methodology}

\subsection{Model assumptions and problem formulation}

As mentioned in the previous section, each pixel in the scene is modeled as response of a BIBO stable, minimum phase LTI measurement channel with unknown  dynamics.  Each LTI channel is assumed to be corrupted by an uncorrelated additive noise with an unknown distribution\footnote{Since the motion of the pixels due to respiration and other sources are additive, it is reasonable to assume the noise to be additive.}. No additional inter-relationship assumptions are required to be made based on geographic proximity between channels although in reality stronger correlation is expected between spatially proximal pixels. Note that the spectral characteristics of the noise is scene specific and hence no distributional assumption is made.\\
We term the periodic physical movements of the chest region caused by flow of air into and out of the respiratory system as the generating signal, a correlate of which (RP) we wish to estimate  using a video stream from a 2D camera consisting of $P$x$Q$ pixels in each frame. Let the generating signal be denoted by $g(t)$. Let the recorded pixel intensity at $i^{th}$ pixel, at a time $t$ be $x_{i}(t)$ and the transfer function of the LTI channel associated with that pixel be $h_{i}(t)$. The noise process associated with that channel shall be denoted by $n_{i}(t)$, with zero mean. Mathematically,

\begin{equation}
x_{i}(t)=h_{i}(t)\otimes g(t)+n_{i}(t)\label{eq:model}
\end{equation}
\[
i=1,2,3....,PQ
\]

\begin{figure}[tbh]
\begin{centering}
\includegraphics[width=3.5in]{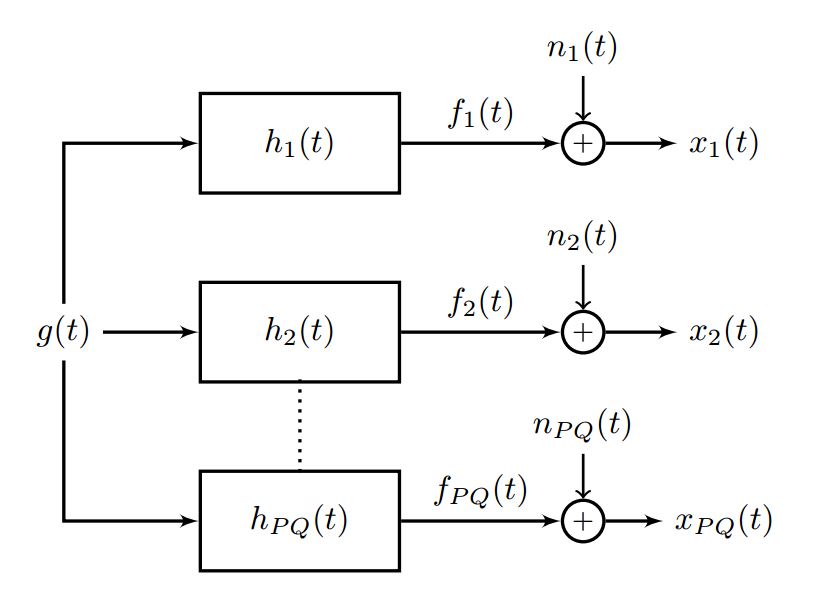}
\par\end{centering}
\caption{Single input multiple output model is assumed for the video. It is
assumed that all the LTI systems (pixels) having different system
responses are driven by the same input $g(t)$. Also, every pixel
has its own additive noise source. \label{fig:SIMO}}
\end{figure}

Here $\otimes$ denotes the convolution operator. Let $|H_{i}(w)|$
and $\angle H_{i}(w)$ denote the magnitude and phase response of
the $i^{th}$ LTI channel. We model the ensemble of $|H_{i}(w)|$
over the variable $i$ as a random process of the variable $w$. Sampling
$|H_{i}(w)|$ at each frequency $w$ yields an IID random variable
indexed by the variable $i$. Also $\angle H_{i}(w$) is assumed to
be sampled from a uniform distribution between \textcolor{black}{$-\pi$
and $\pi$.} The entire video now becomes a single input multiple
output (SIMO) system with the outputs of an ensemble of several LTI
channels being driven by the same signal as depicted in Fig. \ref{fig:SIMO}.

Under this model, the mathematical problem of interest is: Given $x_{1}(t),x_{2}(t),.....,x_{PQ}(t)$,
and that $h_{1}(t),h_{2}(t),......,h_{PQ}(t)$ are unknown, obtain
an estimate of $g(t)$, denoted by $\hat{g}(t)$ which is equal to
$g(t)$ up to an amplitude scaling factor. That is, estimate $\hat{g}(t)=cg(t)$
where $c$ is an arbitrary constant\footnote{Note that this scaling factor is entirely determined by the scene
and is subject specific and can be obtained only through calibration.}. This is intractable in general since no information regarding the
transfer functions of the LTI systems is available. However, we show
that a recovery of $\hat{g}(t)$ is possible if certain assumptions
are made about the characteristics of $g(t)$. Specifically, if $g(t)$
is periodic\footnote{This is a reasonable assumption in the case of respiratory signals
during tidal breathing since they are mostly either periodic or quasi-periodic. }, we show in the subsequent sections that it is possible to recover
$\hat{g}(t)$. To start with, we develop the theory for the case of
a pure tone ($g(t)$ being a single frequency sinusoid) and further
extend it to the case of a general periodic signal. 

\subsection{Solution for a pure-tone case: Lemma - 1}

Let $g(t) = Gsin(w_0t+\theta)$. From the LTI system theory, the output
response of each LTI channel (denoted by $f_{i}(t)$) will be of the
following form: $f_{i}(t) = GF_{i}sin(w_0t + \phi_{i}+\theta)$ where
$F_{i}$ = $|H_{i}(w_{0})|$ and $\phi_{i}=\angle H_{i}(w_{0})$ which
are both unique and unknown for each LTI channel. Now, from Eq. \ref{eq:model},
$x_{i}(t) = GF_{i}sin(w_0t + \phi_{i}+\theta) + n_{i}(t)$. The following
lemma demonstrates the existence of an estimator for $g(t)$.
\begin{lem}
If $g(t)$ is a single frequency sinusoid, the ensemble average of
LTI output responses taken over a membership set $X^{+}$, defined
as \begin{align*}    X^{+} = \left\{\begin{array}{lr}         x_{i}(t) :   |\phi_i | \leq \pi/2\\         \forall i \in \{1,...,PQ \}         \end{array}\right\} \end{align*}asymptotically
converges to a scaled version of the generating signal $g(t)$. Mathematically,
\begin{equation}
\hat{g}(t)=\frac{1}{\overline{X^{+}}}\sum_{i\in X^{+}}x_{i}(t)=cg(t)\label{eq:estimator}
\end{equation}
Here for any set $A$, operator $\overline{A}$  denotes the cardinality
of set $A$.\textbf{ }
\end{lem}
\begin{IEEEproof}
\noindent From Eq. \ref{eq:estimator},
\begin{eqnarray}
\hat{g}(t) & = & \frac{1}{\overline{X^{+}}}\sum_{i\in X^{+}}x_{i}(t)\\
 & = & \frac{1}{\overline{X^{+}}}\sum_{i\in X^{+}}GF_{i}sin(w_{0}t+\phi_{i}+\theta)+n_{i}(t)\label{eq:lemma1sum}
\end{eqnarray}

\noindent For very large $\overline{X^{+}}$, that is  ${\overline {X^{+}}} \to \infty $,
the summation in Eq. \ref{eq:lemma1sum} may be replaced by an expectation
operator (\textcolor{black}{over the joint distribution of random
variables $F_{i}$, $\phi_{i}$ and $n_{i}$, taken over the set corresponding
to $X^{+}$}) at every time instant $t$, by the law of large numbers.
Thus, 
\begin{equation}
\hat{g}(t)=E_{F\phi n}[GF_{i}sin(w_{0}t+\phi_{i}+\theta)+n_{i}(t)]\label{eq:lemma1exp}
\end{equation}
\textcolor{black}{With the assumption of independence between $\phi$,
$F$ and noise and with the linearity of the expectation operator,
Eq. \ref{eq:lemma1exp} may be split as follows, with $E_{F}$, $E_{\phi}$
and $E_{n}$ representing the expectations under the distributions
over the random variables $F$, $\phi$ and $n$, respectively over
the set $X^{+}.$}
\begin{equation}
\hat{g}(t)=GE_{F}[F_{i}]E_{\phi}[sin(wt+\phi_{i}+\theta)]+E_{n}[n_{i}]\label{eq: gcap}
\end{equation}

\noindent \textcolor{black}{By definition, it follows that, over the
set corresponding to $X^{+}$, $\phi\in[\frac{-\pi}{2},\frac{\pi}{2}]$
albeit the support of $\phi$ is $[-\pi,\pi]$. Therefore, in Eq.
\ref{eq: gcap}, 
\begin{eqnarray}
E_{\phi}[sin(wt+\phi_{i}+\theta)] & = & \frac{1}{\pi}\int_{-\frac{\pi}{2}}^{\frac{\pi}{2}}sin(wt+\phi+\theta)d\phi\nonumber \\
 & = & \frac{sin(wt+\theta)}{\pi}\label{eq:expect}
\end{eqnarray}
}

\noindent \textcolor{black}{Thus, with noise process being zero-mean
and from Eq. \ref{eq: gcap} and Eq. \ref{eq:expect}, 
\[
\hat{g}(t)=GE_{F}[F_{i}]\frac{sin(wt+\theta)}{\pi}
\]
}Thus, $\hat{g}(t) \to cGsin(wt+\theta) \qquad\textrm{for}\qquad {\overline {X^{+}}} \to \infty $
where $c=E_F[F_i]/ {\pi}$.
\end{IEEEproof}
Lemma 1 asserts that the ensemble average of the output responses
of a group of LTIs belonging to a set $X^{+}$, converges to the scaled
version of the input, when the input is a pure tone. Such an ensemble
averaging will also reduce the additive noise in the responses. Although
the existence of such a set cannot be guaranteed for every problem,
it can be empirically argued that such a set is very likely to exist
for the cases considered in the current problem of RP estimation.
The rest of this section describes a method for determining the set
$X^{+}$ from a given large set of LTI channel responses. 

\subsection{Finding $X^{+}$ - Quadratic approximation}

Finding $X^{+}$ through a brute-force method (using its definition)
is not feasible because computing phase difference between two signals
corrupted with noise is non-trivial. This is because the phase lag
introduced by each LTI channel and its associated noise level are
unknown. Hence, in this section, we describe an effective method that
would not only serve to estimate g(t) by choosing $X^{+}$, but also
aids in noise reduction. 

It was seen in the previous section that the responses of random LTI
channels to a pure tone excitation signal results in a set of randomly
scaled and shifted sinusoids (see $x_{i}(t)$), which have two degrees
of freedom namely random amplitude scale and phase shift. Intuitively,
such a set of random sinusoids can be mapped isomorphically to a two-dimensional
space in which the arbitrary amplitude and the phase-lag are better
represented. We propose to represent each of the LTI channel responses
using a basis set derived out of standard second degree polynomials.
Following section lists a set of definitions used to formalize the
treatment. 

\subsubsection{Some definitions and notations}

Let $x(t)$ and $y(t)$ represent two finite energy signals in Hilbert
space ($\mathscr{H_{2}}$) and let $x(t)$ be periodic with $w_{0}$
denoting its dominant frequency - the frequency with the highest magnitude
in the Fourier line spectrum of $x(t).$ We define the inner-product
between $x(t)$ and $y(t)$ as in Eq. \ref{eq:inner_product} described
below. 
\begin{equation}
\langle x(t),y(t)\rangle=\frac{w_{0}}{2\pi}\int_{\frac{-\pi}{w_{0}}}^{\frac{\pi}{w_{0}}}x(t)y(t)\,\,dt\label{eq:inner_product}
\end{equation}
 The norm of a signal $x(t)$ is defined as $||x(t)||^{2}=\mathbf{\langle}x(t),x(t)\rangle$.
Note that the value of inner-products and norms are frequency dependent.
We derive a set of basis $\varPsi=\{\psi_{1}(t,w_{0}),\psi_{2}(t,w_{0}),\psi_{3}(t,w_{0})\}$
by orthonormalization of the standard polynomial basis $\Omega=\{1,t,t^{2}\}$
using the Gram-Schmidt procedure. This is to facilitate the easy computation
as evident in the subsequent sections. The individual components of
$\varPsi$ are given by the following equations.

\begin{eqnarray}
\psi_{1}(t,w_{0}) & = & \frac{3\sqrt{5}}{2\pi^{2}}w_{0}^{2}t^{2}-\frac{\sqrt{5}}{2}\label{eq:astar}\\
\psi_{2}(t,w_{0}) & = & \frac{\sqrt{3}}{\pi}w_{0}t\label{eq:bstar}\\
\psi_{3}(t,w_{0}) & = & 1
\end{eqnarray}

\subsubsection{Projection of signals on to the basis \textmd{$\varPsi$}}

Let $Q(t)$ as denoted in Eq. \ref{eq:qdef} represent the span of
the basis $\varPsi$ in $\mathcal{\mathtt{\mathscr{R^{3}}}}$. 
\begin{equation}
Q(t)\triangleq a\psi_{1}(t,w_{0})+b\psi_{2}(t,w_{0})+c\psi(t,w_{0})\label{eq:qdef}
\end{equation}
where $a,b$ and $c$ are real numbers. The optimal coefficients $\{a^{*},b^{*},c^{*}\}$
representing the best fit of a signal $s(t)$ in the span of $\varPsi$
are obtained by solving the optimization problem in Eq. \ref{eq:optim}.

\begin{equation}
\{a^{*},b^{*},c^{*}\}=argmin_{\{a,b,c\}}||Q(t)-s(t)||^{2}\label{eq:optim}
\end{equation}

Denoting the error function as $E$, the optimal solution to the problem
in Eq. \ref{eq:optim} is obtained by simultaneously solving $\partial E/\partial a=0,\,\,\,\partial E/\partial b=0$
and $\partial E/\partial c=0$ which yields the following equations:

\begin{equation}
\mathbb{V}\begin{bmatrix}a^{*}\\
b^{*}\\
c^{*}
\end{bmatrix}=\begin{bmatrix}\langle s(t),\psi_{1}(t,w_{0})\rangle\\
\langle s(t),\psi_{2}(t,w_{0})\rangle\\
\langle s(t),\psi_{3}(t,w_{0})\rangle
\end{bmatrix}\label{eq:vander}
\end{equation}

\noindent where $\mathbb{V}_{i,j}=\langle\psi_{i},\psi_{j}\rangle$.
Noting that under the defined basis, $\mathbb{V}$ is an identity
matrix, from Eq. \ref{eq:vander}, 

\noindent 
\begin{eqnarray}
a^{*} & = & \langle s(t),\psi_{1}(t,w_{0})\rangle\nonumber \\
b^{*} & = & \langle s(t),\psi_{2}(t,w_{0})\rangle\\
c^{*} & = & \langle s(t),\psi_{3}(t,w_{0})\rangle\nonumber 
\end{eqnarray}
 It is to be observed that $c^{*}$ is the mean of the signal $s(t)$
which can be forced to zero if all signals are enforced to be zero-mean.
In the rest of the paper, we omit $c^{*}$ since we only consider
zero-mean signals. Representation of each of the output responses
of the aforementioned random LTI systems ($x_{i}(t)$) using the quadratic
approximation may be summarized using the following lemma. 
\begin{lem}
Let $s_{i}(t)=sin(w_{0}t+\phi_{i}+\theta)$ and let $\left\{ a_{i}^{*},b_{i}^{*}\right\} $
represent the solution for the optimization problem in Eq. \ref{eq:optim}.
Let$\left\{ \mathscr{A},\mathcal{\mathscr{B}}\right\} $denote the
points spanned by all the solutions for $\phi_{i}\sim\mathscr{\mathbb{U}}\left[-\pi,\pi\right]$,
then \textup{$\left\{ \mathscr{A},\mathcal{\mathscr{B}}\right\} $
}lies on the periphery of an ellipse in the solution space. Here $\mathbb{U}$
denotes a uniform distribution.
\end{lem}
\begin{enumerate}
\item \textbf{Corollary 1: }\textit{The major axis of the ellipse is the
line corresponding to $\phi=-\theta$.}\textbf{\textit{ }}
\item \textbf{Corollary 2:}\textbf{\textit{ }}\textit{If $s_{i}(t)=$$\alpha_{i}sin(wt+\phi_{i}+\theta)$
with $\alpha_{i}\leq\alpha_{max}$ the solution space will lead to
a filled ellipse with the distance of a point from center of the ellipse
being proportional to the corresponding amplitude. }
\end{enumerate}
\begin{IEEEproof}
\noindent It is easy to see that

\noindent 
\begin{eqnarray}
a_{i}^{*} & = & \langle sin(w_{0}t+\phi_{i}+\theta),\psi_{1}(t,w_{0})\rangle\nonumber \\
 & = & \frac{3\sqrt{5}}{\pi^{2}}sin(\phi_{i}+\theta)
\end{eqnarray}
And similarly,

\noindent 
\begin{eqnarray}
b_{i}^{*} & = & \langle sin(w_{0}t+\phi_{i}+\theta),\psi_{2}(t,w_{0})\rangle\nonumber \\
 & = & \frac{\sqrt{3}}{\pi}cos(\phi_{i}+\theta)
\end{eqnarray}
for $\phi_{i}\sim\mathscr{\mathbb{U}}\left[-\pi,\pi\right]$, equations
for $a_{i}^{*}$and $b_{i}^{*}$ represent the parametric form for
an ellipse in the space of $a$ and $b$ whose major axis corresponds
to the line $\phi_{i}+\theta=0$ or $\phi_{i}=-\theta$. Also, for
\textit{$s_{i}(t)=$$\alpha_{i}sin(wt+\phi_{i}+\theta)$, 
\begin{equation}
a_{i}^{*}=\alpha_{i}\frac{3\sqrt{5}}{\pi^{2}}sin(\phi_{i}+\theta)
\end{equation}
 }and\textit{ 
\begin{equation}
b_{i}^{*}=\alpha_{i}\frac{\sqrt{3}}{\pi}cos(\phi_{i}+\theta)
\end{equation}
 }which are concentric ellipses for different $\alpha_{i}$ which
are bounded within the ellipse for \textit{$\alpha_{i}=\alpha_{max}$.}
\end{IEEEproof}

\subsection{Extension to a general periodic signal}

The discussion so far has only dealt with a pure-tone albeit in practice
the signals that are encountered will have multiple harmonics. In
this section, the extensions of Lemmas 1 and 2 to the case of general
periodic signal will be discussed. Let 
\begin{equation}
g(t)=\sum_{k=1}^{\infty}G_{k}sin(w_{k}t+\theta)
\end{equation}
be the generating signal of interest\footnote{Without the loss of generality it can be assumed that the phase term
$\theta$ is constant for all harmonics. This is because, any periodic
signal can be decomposed in to its even and odd periodic components,
each of which has a constant phase term for all harmonics.}. In the rest of the paper, for simplicity of analysis we restrict
the Fourier representation of $g(t)$ to $N$ significant harmonics
each at $w_{k}$. 

\subsubsection{Estimator for a general periodic signal}

From Sections II A and B for this case of $g(t)$, 
\begin{equation}
x_{i}(t)=\sum_{k=1}^{N}G_{k}F_{i}(w_{k})sin(w_{k}t+\phi_{i}(w_{k})+\theta)+n_{i}(t)
\end{equation}
 where $F_{i}(w_{k})=|H_{i}(w_{k})|$ and $\phi_{i}(w_{k})=\angle H_{i}(w_{k})$.
With these definitions, Lemma 3 describes the condition for the previously
defined estimator, that is,  $\hat{g}(t) = \frac{1}{\overline {X^{+}}} \sum_{i \in X^{+}} x_{i}(t)$
to converge to $cg(t)$, in addition to that laid in Lemma 1. 
\begin{lem}
$\hat{g}(t)$ converges to $g(t)$ for ${\overline {X^{+}}} \to \infty$
if ${E_{F}[F_i(w_k)]} = constant \,\,\,, \forall k$.
\end{lem}
\begin{IEEEproof}
\noindent \begin{flushleft}
By definition,\begin{align*} \hat{g}(t)=\sum\limits _{i\in X^{+}}\sum\limits _{k=1}^{N}G_kF_{i}(w_{k})sin(w_{k}t+\phi_{i}(w_{k})+\theta)+n_{i}(t)\end{align*}Now,
if the phase delay offered by all the channels is assumed to be a
constant at all $w_{k}$\footnote{\textcolor{black}{It is known that CMOS image sensors that are used
in most of the cameras have a very wide frequency response, often
up to 1 MHz and can achieve very high frame rates \cite{johnston20112d,fossum1997cmos,el2009cmos}.
The frame rate necessary for applications such as the one in this
paper, does not exceed 30 fps which represent signals up to 15 Hz.
Since every pixel is modeled as an LTI channel that has a bandwidth
of order of MHz, it is reasonable to assume a constant phase delay
for each channel (CMOS sensor) over the small frequency range of interest
(a few Hz).}}, $\phi_{i}(w_{k})$ will be independent of $w_{k}$ which can be
represented as $\phi_{i}$. As in Lemma 1, for very large $\overline{X^{+}}$,
that is  $\overline{X^{+}} \to \infty $, the outer summation in the
definition of $\hat{g}(t)$ can be replaced by an expectation operator
by the law of large numbers. Thus, 
\par\end{flushleft}
\begin{eqnarray}
\hat{g}(t) & = & E[\sum_{k=1}^{N}G_{k}F_{i}(w_{k})sin(w_{k}t+\phi_{i}+\theta)+n_{i}(t)]\nonumber \\
 & = & \sum_{k=1}^{N}G_{k}E_{F}[F_{i}(w_{k})]E_{\phi}[sin(w_{k}t+\phi_{i}+\theta)]+E_{n}[n_{i}]\nonumber \\
 & = & \sum_{k=1}^{N}G_{k}E_{F}[F_{i}(w_{k})]\frac{sin(w_{k}t+\theta)}{\pi}\label{eq:lemma3conv}
\end{eqnarray}
Note that in the above expressions as in the case with Lemma 1
\begin{equation}
E_{\phi}[sin(w_{k}t+\phi_{i}+\theta)]=\frac{sin(w_{k}t+\theta)}{\pi}
\end{equation}
 over set $X^{+}$. From Eq. \ref{eq:lemma3conv}, 
\[
\hat{g}(t)\to c\sum\limits _{k=1}^{N}G_{k}sin(w_{k}t+\theta)
\]
for $\overline{X^{+}}\to\infty$ if ${E_{F}[F_i(w_k)]} = c = constant$.
\end{IEEEproof}
Lemma 3 along with Lemma 1, asserts that $X^{+}$ should be chosen
such that the phase-lags introduced by each of the LTI-channels in
the set $X^{+}$ should be within $\pi/2$ radians of $\theta$ and
${E_{F}[F_i(w_k)]} = constant$. In the subsequent section, we show
that projection of the signals on the aforementioned quadratic basis
aids to select points satisfying both the conditions. 

\subsubsection{Quadratic basis projection for general periodic signal}

Let the output response of an $i^{th}$ random LTI system described
in Sec. II B, when excited by a periodic signal $g(t)$ described
by Eq. \ref{eq:gg} 
\begin{equation}
g(t)=\ensuremath{\sum\limits _{k=1}^{N}G_{k}sin(w_{k}t+\theta)}\label{eq:gg}
\end{equation}
where $w_{k}=kw_{0}$, be represented by $f_{i}(t)$, given by Eq.
\ref{eq:fi}. 
\begin{equation}
f_{i}(t)=\ensuremath{\sum\limits _{k=1}^{N}F_{i}(w_{k})G_{k}sin(w_{k}t+\phi_{i}+\theta)}\label{eq:fi}
\end{equation}

\begin{figure}[tbh]
\begin{centering}
\includegraphics[width=3.4in]{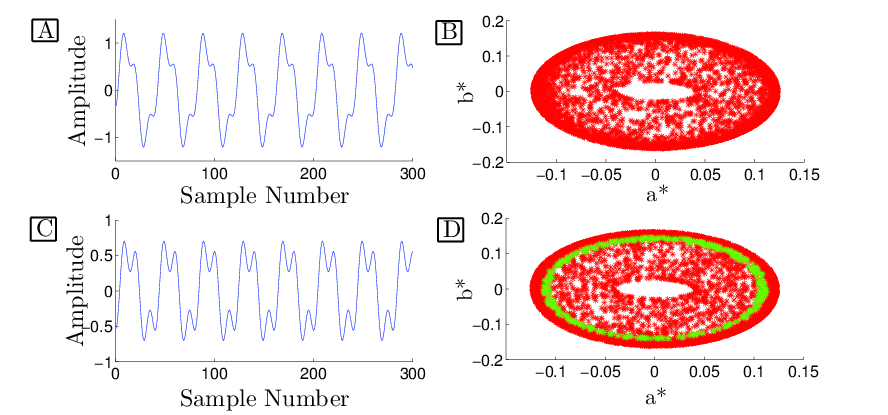}
\par\end{centering}
\caption{Demonstration of Lemma 4 and its implications : (A) and (B) respectively
depict the generating signal ($g(t)=sin(50\pi t)+\frac{1}{3}sin(150\pi t)$)
and the elliptical disk generated from quadratic fit coefficients
corresponding to the outputs of several random LTI channels. (C) depicts
the output of a random LTI channel with a given magnitude response,
when excited by $g(t).$ (D) is the disk in (B), with the elliptical
ring corresponding to set of LTI channels with the magnitude response
used to obtain the signal (C), marked by green dots. \label{fig:lemma4}}
\end{figure}

\begin{lem}
The ensemble of the quadratic fit coefficients $\{a_{i}^{*},b_{i}^{*}\}$
for \textup{$f_{i}(t)=\ensuremath{\sum\limits _{k=1}^{N}F_{i}(w_{k})G_{k}sin(w_{k}t+\phi_{i}+\theta)}$}
with $\phi_{i}\sim\mathscr{\mathbb{U}}[0,2\pi]$, defines a filled
parametric elliptical disk in the coefficient space. 
\end{lem}
\begin{IEEEproof}
\noindent \begin{flushleft}
From Section, II.C, we know that for any signal $f_{i}(t)$, the least-square
quadratic fit coefficients $\{a_{i}^{*},b_{i}^{*}\}$ on a basis set
$\varPsi$ are given by $a_{i}^{*}=\langle f_{i}(t),\psi_{1}(t,w_{0})\rangle$
and $b_{i}^{*}=\langle f_{i}(t),\psi_{2}(t,w_{0})\rangle$. From Eq.
\ref{eq:astar} and \ref{eq:bstar},
\begin{eqnarray}
a_{i}^{*} & = & \ensuremath{\sum_{k=1}^{N}F_{i}(w_{k})G_{k}\langle sin(w_{k}t+\phi_{i}+\theta),\psi_{1}(t,w_{0})\rangle}\\
 & = & \sum_{k=1}^{N}F_{i}(w_{k})G_{k}I_{i}^{k}\label{eq:innerpr}
\end{eqnarray}
where $  I^k_i={\langle sin(w_{k}t+\phi_{i}+\theta), \psi_{1}(t,w_{0}) \rangle}$,
the $k^{th}$ inner product term in Eq. \ref{eq:innerpr}. Noting
that $w_{k}=kw_{0}$, 
\begin{align}
I_{i}^{k} & =\ensuremath{\frac{3\sqrt{5}}{\pi^{2}}sin(\phi_{i}+\theta)\frac{(-1)^{k}}{k^{2}}}\label{eq:Ik}
\end{align}
\par\end{flushleft}
\noindent Following the linearity of the inner products and Eq. \ref{eq:innerpr}
and \ref{eq:Ik},
\begin{align}
a_{i}^{*} & =\ensuremath{\frac{3\sqrt{5}}{\pi^{2}}sin(\phi_{i}+\theta)\sum_{k=1}^{N}F_{i}(w_{k})G_{k}\frac{(-1)^{k}}{k^{2}}}\label{eq:astarg}
\end{align}
Similarly,
\begin{align}
b_{i}^{*} & =\ensuremath{\frac{\sqrt{3}}{\pi}cos(\phi_{i}+\theta)\sum_{k=1}^{N}F_{i}(w_{k})G_{k}\frac{(-1)^{k+1}}{k}}\label{eq:bstarg}
\end{align}

\noindent Since the summation terms in Eq. \ref{eq:astarg} and \ref{eq:bstarg}
converge to a finite number, $\{a_{i}^{*},b_{i}^{*}\}$ define a parametric
ellipse for $\phi_{i}\sim\mathscr{\mathbb{U}}[-\pi,\pi]$. Also, depending
upon the values of the product terms $F_{i}(w_{k})G_{k}$, they converge
to a different number, leading to a filled ellipse.
\end{IEEEproof}
\begin{flushleft}
The following are some of the major implications of Lemma 4 which
are to be noted. 
\par\end{flushleft}
\begin{enumerate}
\item Every point on the filled ellipse corresponds to an LTI channel with
a certain magnitude and phase response.
\item The major axis corresponds to the that LTI channel with a phase response
$\phi_{i}=-\theta$ . LTI channels with all other phase shifts ($\phi_{i}$)
are symmetrically and uniformly distributed around the major axis.
\item A set of LTI channels with the same magnitude response but different
phase response correspond to points lying on an elliptical ring inside
the disk. This is evident from Eq. \ref{eq:astarg} and \ref{eq:bstarg},
where, for the LTI channels with same magnitude response, $F_{i}(w_{k})$
is independent of $i$. Further, since $G_{k}$ is fixed, $\{a_{i}^{*},b_{i}^{*}\}$
for such a set defines an ellipse with a fixed length major and minor
axis. 
\item If the generating signal is assumed to be of a low-pass nature, that
is, $|G_{k}|>|G_{k+1}|$\footnote{Most biomedical signals, including the respiratory pattern show decreasing spectral magnitude.},
the points closer to the periphery of the disk, correspond to the
LTI channels that emphasize fundamental frequency the most, over the
harmonics. This is because, in this case, the summation terms in Eq.
\ref{eq:astarg} and \ref{eq:bstarg} are monotonically decreasing
series with alternating sign.
\item The points that are farther away from the periphery of the disk, correspond
to the LTI channels  that attenuate the fundamental frequency while
emphasizing the higher harmonics. 
\end{enumerate}
Fig. \ref{fig:lemma4} demonstrates Lemma 4 and some of its implications.

\subsection{Impact of noise on the coefficients}

The model proposed in Sec. II A, involves an additive noise component
associated with each pixel (LTI channel) that has not been considered
in all the analysis so far. In this section, the impact of additive
noise on the coefficients obtained from quadratic polynomial fitting
is discussed. 

For a periodic excitation signal $g(t)=\ensuremath{\sum\limits _{k=1}^{N}G_{k}sin(w_{k}t+\theta)}$,
from Sec. II A, we have the response of each individual LTI system,
\begin{equation}
x_{i}(t)=f_{i}(t)+n_{i}(t)
\end{equation}
 with 
\begin{equation}
f_{i}(t)=\ensuremath{\sum\limits _{k=1}^{N}F_{i}(w_{k})G_{k}sin(w_{k}t+\phi_{i}+\theta)}
\end{equation}
From Sec. II.C.2, we know that the quadratic coefficients for the
signal $x_{i}(t)$ are given by $\hat{a}_{i}=\langle x_{i}(t),\psi_{1}(t,w_{0})\rangle$,
$\hat{b}_{i}=\langle x_{i}(t),\psi_{2}(t,w_{0})\rangle$ and because
we are working with zero-mean signals, $c_{i}^{*}=0$. Since the inner
products are linear, 
\begin{eqnarray}
\hat{a}_{i} & = & \langle f_{i}(t),\psi_{1}(t,w_{0})\rangle+\langle n_{i}(t),\psi_{1}(t,w_{0})\rangle\\
\hat{b}_{i} & = & \langle f_{i}(t),\psi_{2}(t,w_{0})\rangle+\langle n_{i}(t),\psi_{2}(t,w_{0})\rangle
\end{eqnarray}
  Let $\langle f_{i}(t),\psi_{1}(t,w_{0})\rangle=a_{i}^{*}$ and $\langle f_{i}(t),\psi_{2}(t,w_{0})\rangle=b_{i}^{*}$
represent the solution for the no-noise case. Given the aforementioned
definitions, the objective is to relate $\{\hat{a}_{i}$,$\hat{b}_{i}\}$
to $\{a_{i}^{*},b_{i}^{*}\}$. We have,

\begin{eqnarray}
\hat{a}_{i} & = & a_{i}^{*}+\langle n_{i}(t),\psi_{1}(t,w_{0})\rangle\label{eq:acap}\\
\hat{b}_{i} & = & b_{i}^{*}+\langle n_{i}(t),\psi_{2}(t,w_{0})\rangle\label{eq:bcap}
\end{eqnarray}

From Cauchy-Shwartz inequality, 
\begin{eqnarray}
-\sigma & \leq & \langle n_{i}(t),\psi_{1}(t,w_{0})\rangle\leq\sigma\label{eq:cs1}\\
-\sigma & \leq & \langle n_{i}(t),\psi_{2}(t,w_{0})\rangle\leq\sigma\label{eq:cs2}
\end{eqnarray}
 where $||n_{i}(t)||^{2}=\sigma^{2}$ and $||\psi_{i}||^{2}=1$, by
definition. Thus from Eq. \ref{eq:acap}, \ref{eq:bcap}, \ref{eq:cs1}
and \ref{eq:cs2},

\begin{eqnarray}
a_{i}^{*}-\sigma\leq & \hat{a}_{i} & \ensuremath{\leq a_{i}^{*}+\sigma}\label{eq:apert}\\
b_{i}^{*}-\sigma\leq & \hat{b}_{i} & \leq b_{i}^{*}+\sigma\label{eq:bpert}
\end{eqnarray}

\begin{figure}[tbh]
\begin{centering}
\includegraphics[width=3.4in]{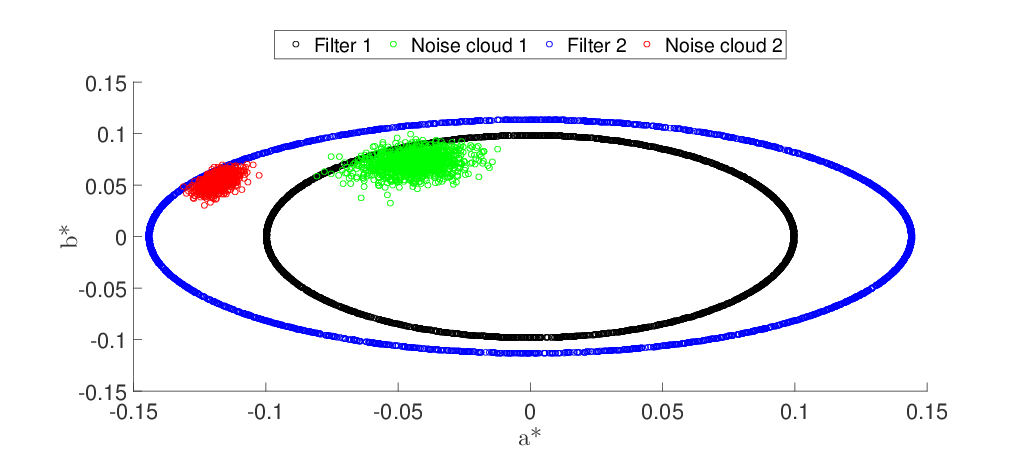}
\par\end{centering}
\caption{Illustration of Lemma 5 - Ellipses in blue and black, respectively,
represent the quadratic coefficient space for the output of two sets
of LTI channels with magnitude response $[F_{1}(40\pi),F_{1}(80\pi)]=\{0.8,0.1\}$
and$[F_{2}(40\pi),F_{2}(80\pi)]=\{0.2,0.8\}$ and phase shifts defined
by$\phi_{i}\sim\mathscr{\mathbb{U}}[-\pi,\pi]$ driven by a same generating
signal $g(t)=sin(40\pi t)+0.5sin(80\pi t)$. Thousand samples of Gaussian
noise process with $\sigma=0.5$ are generated and added to the output
of one LTI channel with a fixed phase taken from each set. The quadratic
fits for the noisy signals are shown using red and green dots. It
can be seen that for the same amount of noise power, the area of cloud
of perturbation is lower for the LTI channel that lies exterior which
corresponds to the LTI channel emphasizing the fundamental frequency.
\label{fig:noisecloud}}
\end{figure}

From Eq. \ref{eq:apert} and \ref{eq:bpert}, it can be inferred that
with the addition of noise, $\{a_{i}^{*},b_{i}^{*}\}$ gets perturbed
within a cloud bounded by $|\sigma|$.Since there is no natural comparative
bound of the relative magnitudes of noise and coefficients, nothing
can be inferred regarding the relation between the position of a given
point in the coefficient space and the quality of the signal. However,
useful insights can be obtained if all the signals are normalized
(forced to be unit norm) prior to quadratic fitting. Let the signal-to-noise-ratio
(SNR) corresponding to $i^{th}$ LTI channel denoted by $\rho_{i}$,
be defined as $\rho_{i}\triangleq||f_{i}(t)||/||n_{i}(t)||.$ With
these notations, the following Lemma relates $\{\hat{a}_{i},\hat{b}_{i}\}$,
$\{a_{i}^{*},b_{i}^{*}\}$ and $\rho_{i}.$ 
\begin{lem}
When random noise $n_{i}(t)$ is added to $f_{i}(t)$ to yield $x_{i}(t)$,
the quadratic coefficients \textup{$\{a_{i}^{*},b_{i}^{*}\}$} corresponding
to normalized $f_{i}(t)$ get scaled by a factor less than unity and
perturbed within a cloud whose area is inversely proportional to $\rho_{i}^{2}$
to yield the quadratic coefficients corresponding to the noisy signal.\textup{ }
\end{lem}
\begin{IEEEproof}
\noindent Let $x_{i}(t)$ be forced to have unit norm before quadratic
approximation to yield $\tilde{x}_{i}(t)$ $=x_{i}(t)/||x_{i}(t)||$.
By definition, 
\begin{eqnarray}
||x_{i}(t)||^{2} & = & ||f_{i}(t)+n_{i}(t)||^{2}\nonumber \\
 & = & ||f_{i}(t)||^{2}+||n_{i}(t)||^{2}\\
 & = & \sigma^{2}(\rho_{i}^{2}+1)\nonumber 
\end{eqnarray}
 because $\langle f_{i}(t),n_{i}(t)\rangle=0$. Note that
\begin{eqnarray}
a_{i}^{*} & = & \langle f_{i}(t)/||f_{i}(t)||,\psi_{1}(t,w_{0})\rangle\\
 & \Rightarrow & |f_{i}(t)||a_{i}^{*}=\langle f_{i}(t),\psi_{1}(t,w_{0})\rangle
\end{eqnarray}

\begin{eqnarray}
b_{i}^{*} & = & \langle f_{i}(t)/||f_{i}(t)||,\psi_{2}(t,w_{0})\rangle\\
 & \Rightarrow & ||f_{i}(t)||b_{i}^{*}=\langle f_{i}(t),\psi_{2}(t,w_{0})\rangle
\end{eqnarray}

\noindent Let $\{\hat{a}_{i},\hat{b}_{i}\}$ denote the quadratic
coefficients for $\tilde{x}_{i}(t)$. From Lemma 2, we have 
\begin{eqnarray}
\hat{a}_{i} & = & \frac{||f_{i}(t)||a_{i}^{*}}{||x_{i}(t)||}+\frac{\langle n_{i}(t),\psi_{1}(t,w_{0})\rangle}{||x_{i}(t)||}\label{eq:acap1}\\
\hat{b}_{i} & = & \frac{||f_{i}(t)||b_{i}^{*}}{||x_{i}(t)||}+\frac{\langle n_{i}(t),\psi_{2}(t,w_{0})\rangle}{||x_{i}(t)||}\label{eq:bcap1}
\end{eqnarray}
 From Eq. \ref{eq:acap1}, \ref{eq:bcap1}, \ref{eq:cs1} and \ref{eq:cs2},

\begin{align}
\frac{||f_{i}(t)||a_{i}^{*}}{||x_{i}(t)||}-\frac{\sigma}{||x_{i}(t)||}\leq & \hat{a}_{i}\leq\frac{||f_{i}(t)||a_{i}^{*}}{||x_{i}(t)||}+\frac{\sigma}{||x_{i}(t)||}\label{eq:ahat1}\\
\frac{||f_{i}(t)||b_{i}^{*}}{||x_{i}(t)||}-\frac{\sigma}{||x_{i}(t)||}\leq & \hat{b}_{i}\leq\frac{||f_{i}(t)||b_{i}^{*}}{||x_{i}(t)||}+\frac{\sigma}{||x_{i}(t)||}\label{eq:bhat1}
\end{align}
Using the definition of $||x_{i}(t)||^{2}$ in Eq. \ref{eq:ahat1}
and \ref{eq:bhat1},
\begin{align}
\frac{\rho_{i}a_{i}^{*}}{\sqrt{\rho_{i}^{2}+1}}-\frac{1}{\sqrt{\rho_{i}^{2}+1}}\leq & \hat{a}_{i}\leq\frac{\rho_{i}a_{i}^{*}}{\sqrt{\rho_{i}^{2}+1}}+\frac{1}{\sqrt{\rho_{i}^{2}+1}}\label{eq:ahat2}\\
\frac{\rho_{i}b_{i}^{*}}{\sqrt{\rho_{i}^{2}+1}}-\frac{1}{\sqrt{\rho_{i}^{2}+1}}\leq & \hat{b}_{i}\leq\frac{\rho_{i}b_{i}^{*}}{\sqrt{\rho_{i}^{2}+1}}-\frac{1}{\sqrt{\rho_{i}^{2}+1}}\label{eq:bhat2}
\end{align}

From Eq. \ref{eq:ahat2} and \ref{eq:bhat2}, since ${\rho_i} / {\sqrt{{\rho_i}^2+1}} \leq 1 $
the factor scaling $a_{i}^{*}$ is less than unity and the area of
the cloud of perturbation is ${({{{\rho_i}^2+1}})}^{-1}$.
\end{IEEEproof}
One of the primary implications of Lemma 5 is that for a given amount
of noise power $\sigma$, the signals having a higher $||f_{i}(t)||$
will have a higher SNR $\rho_{i}$. From Lemma 4, it is known that,
for a low-pass signal, the LTI channels that emphasize the fundamental
frequency over the others will have a higher $||f_{i}(t)||$ and hence
a higher SNR. This implies that such LTI channels (mapping to points
closer to the periphery of the elliptical disk defined in Lemma 4)
are likely to be perturbed the least and have a smaller cloud of perturbation.This
fact is illustrated in Fig. \ref{fig:noisecloud} with an example. 

\subsection{Selection of an optimum membership set}

\subsubsection{\textcolor{black}{Criteria for optimality}}

All the discussions so far are aimed towards selecting a membership
set $X^{+}$over which an ensemble average has to be computed to get
an estimate of the generating signal. In this section, we consolidate
all the criteria that are developed in previous sections to select
$X^{+}$ and map them to geometrical locations on the elliptical disks
obtained through quadratic fitting. \textcolor{black}{The theory developed
so far lays the following criteria for optimality of the selected
$X^{+}$.}
\begin{enumerate}
\item There need to be enough channels in $X^{+}$ so that expectations
in all the Lemmas are well approximated by summations. 
\item Lemma 3 demands that for accurate estimation of $g(t)$, $E_{F}(F(w_{k}))=constant.$
Note also that there can be multiple LTI channels having identical
magnitude responses. This implies that, to satisfy condition laid
in Lemma 3, every type of LTI channel magnitude response should receive
equal weightage in $X^{+}$.
\item The phase-lags introduced by each channel in $X^{+}$ must be within
$\pi/2$ radians of the phase of the generating signal $\theta$. 
\item With the addition of noise the channels selected should be such that
the phase distortion caused by the noise should not violate condition
3. 
\end{enumerate}
Any half elliptical arc about the major axis of the disk will ensure
that condition 3 is satisfied. From Lemma 4 the points on the periphery
of the ellipse should be complemented by points on the interior to
satisfy condition 2. An ideal choice which will ensure that both conditions
1 and 2 are met would be to take the entire half ellipse. This is
because $F(w_{k})$ is modeled as an IID random variable whose expectation
would converge to a fixed number and LTI channels with same magnitude
response are represented along the radial arcs of the disk for $\phi_{i}\sim\mathbb{U}[-\pi,\pi${]}.
However, with noise, inclusion of points that are interior on the
disk will distort the morphology of the estimated generating signal
by violating condition 3. Hence, we propose to select the channels
by aggregating along half elliptical arcs starting from the periphery.
However, since the points close to the periphery emphasize the fundamental
frequency and the interior points emphasize the higher harmonics,
as we move inwards, there is a trade-off between getting better estimates
of the higher harmonics and reducing the impact of noise on the obtained
estimate \footnote{Note that the choice of periphery as the starting point is to make sure that the fundamental frequency is not lost.}. 

\subsubsection{\textcolor{black}{Choice of $X^{+}$}}

\textcolor{black}{The optimal choice of $X^{+}$ should adhere to
all the criteria listed above and also to handle the aforementioned
trade-off. We propose to select $X^{+}$ as a set difference between
two sets of points on the feature elliptical disk. }

\textcolor{black}{Let $\{a_{i},b_{i}\}$ represent a point on the
ellipse, $i\in\{1,...,PQ\}$. Let $A=max(a_{i})$ and $B=max(b_{i})$.
Define two sets $D_{O}$ and $D_{I}$ that would correspond to certain
regions within the disk. }

\begin{equation}
\begin{aligned} 
D_{O}= {}  \{ &a_{i},b_{i}:a_{i}\leq Asin(\phi_{i}+\theta),\\ 
& b_{i}\leq Bcos(\phi_{i}+\theta),\\ 
& \forall i:|\phi_{i}|\leq\frac{\pi}{2} \}  \end{aligned} \end{equation}

\begin{equation}
\begin{aligned} 
D_{I}= {}  \{ &a_{i},b_{i}:a_{i}\leq r_eAsin(\phi_{i}+\theta),\\ 
& b_{i}\leq r_eBcos(\phi_{i}+\theta),\\ 
& \forall i:|\phi_{i}|\leq\frac{\pi}{2} \}  \end{aligned} \end{equation} 

where $0<r_{e}\leq1$ denotes the radius of exclusion. 

\textcolor{black}{
\begin{equation}
X^{+}=D_{O}\backslash D_{I}
\end{equation}
}

\begin{figure}[tbh]
\begin{centering}
\textcolor{black}{\includegraphics[width=3.7in]{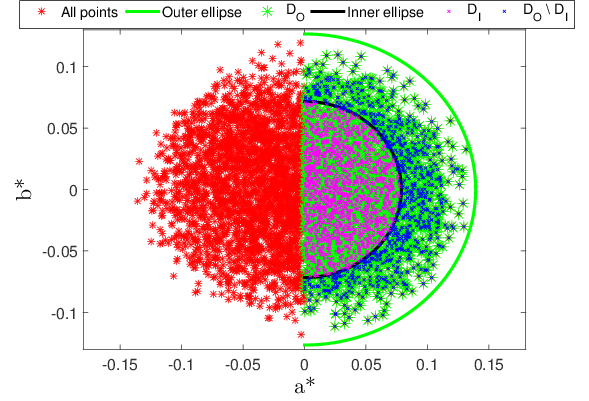}}
\par\end{centering}
\textcolor{black}{\caption{\textcolor{black}{Illustration of procedure to select $X^{+}$. Every
point on the elliptical disk (red crossed points) correspond to one
LTI channel. Green solid line is the outermost half ellipse which
defines the set $D_{O}$ (all the points on one half of the disk within
the outermost ellipse, marked by green star points). A choice of radius
of exclusion $r_{e}$, defines an inner ellipse (marked by black solid
line) which specifies the set $D_{I}$ (the points on one half of
the disk within the inner ellipse, marked by pink crossed points).
Once the sets $D_{O}$ and $D_{I}$ are selected, $X^{+}$ are taken
as the points corresponding to the set difference between $D_{O}$
and $D_{I}$ (marked by blue crossed points). Note that the length
of the axes of the inner ellipse depends upon the value of radius
of exclusion $r_{e}$.\label{fig:Illustration-of-procedure}}}
}
\end{figure}

\begin{figure*}[tbh]
\centering{}\textcolor{black}{\includegraphics[width=6in]{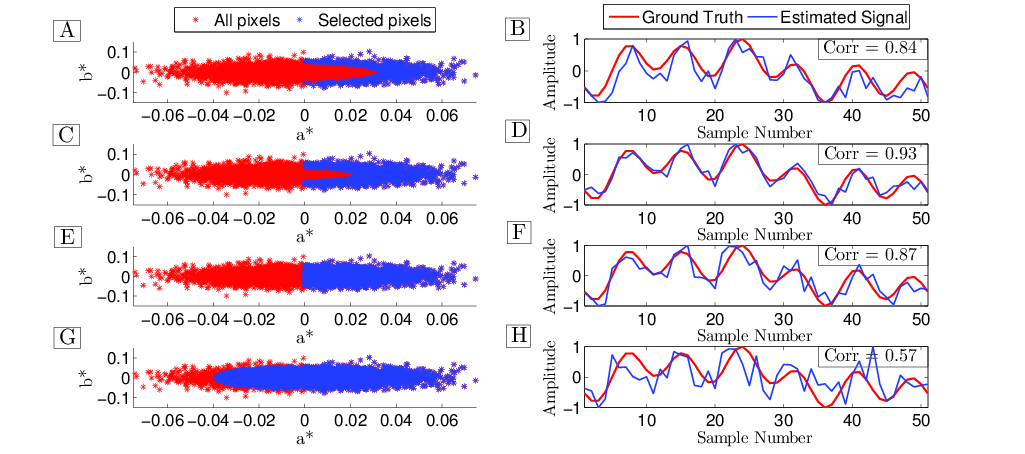}\caption{\textcolor{black}{Illustration of the trade-off between inclusion
of more LTI channels (and thus getting a better estimate of higher
harmonics) in $X^{+}$ and reducing the impact of noise on the estimate:
A generating periodic signal with three harmonics is passed through
random LTI systems with additive white Gaussian noise as discussed
in Sec. II A. (A), (C), (E) and (G) represent the ellipses corresponding
to the quadratic fits of the responses of all the random LTI systems
to the generating signal, with the LTI channels selected (different
set in different cases) to be in $X^{+}$ marked in blue. (B), (D),
(F) and (H) show one period of the corresponding estimated signals
(blue line) along with the generating signal (red line). It can be
seen that the estimate corresponding to the entire half ellipse (E
and F) and one having points close to periphery (A and B) do not agree
as much as the one with an optimized parameter (C and D). In other
words, increase in $r_{e}$ (and thus increasing $\overline{D_{I}}$
) characterizes an increase in the amount of noise in the estimate.
Also, selecting a set of points spanning more than one half of the
disk (G and H, violating the definition of $X^{+}$) greatly distorts
the signal as mentioned in points 3 and 4 of section II. F. These
facts are also corroborated with the value of the normalized correlation
coefficients between the estimates and the ground truth as shown in
the corresponding figures. \label{fig:thresh}}}
}
\end{figure*}

\textcolor{black}{Notice that $D_{O}$ is set of all points within
half the disk along the major axis pointing towards $\phi_{i}=\theta$,
which can also be viewed as set of all points within a half ellipse
whose major axis is $A$. $D_{I}$ is a subset of $D_{O}$ consisting
of points within an ellipse with the length of major and minor axes
as $r_{e}A$ and $r_{e}B$, respectively, where $r_{e}\leq1$ is a
parameter of the method termed the radius of exclusion. A goodness-of-estimation
(GoE) measure is defined (in the subsequent section) which would determine
the `best' choice of this parameter for the selection of $X^{+}$.
Figure \ref{fig:Illustration-of-procedure} illustrates the aforementioned
procedure for selection of $X^{+}$ using an example elliptical disk
in the feature space. }
\textcolor{red}{{} }\textcolor{black}{While an arbitrary value of $r_{e}$
is used in Fig. \ref{fig:Illustration-of-procedure} for an illustration
purpose, Fig. \ref{fig:thresh} illustrates the trade-off between
inclusion of more LTI channels (and thus getting a better estimate
of higher harmonics) in $X^{+}$ and reducing the impact of noise
on the estimate using different values of $r_{e}$ with the corresponding
time-domain signals. It is noteworthy that increase in $r_{e}$ (and
thus increasing $\overline{D_{I}}$ ) characterizes increase in noise
where as it estimates higher harmonics better and vice-versa.}\textcolor{red}{{}
}\textcolor{black}{}

\subsection{Details of implementation for RP estimation}

Implementing the algorithm described in the previous sections for
the current problem of respiration pattern estimation involves computation
of certain parameters outside of the theoretical description, which
will be detailed in this section. Note that for a given video, individual
time series (refer to as pixel time series, PTS) corresponding to
pixel intensity values of each spatial location for a given duration
forms the LTI responses. Also, RP is the generating signal which we
seek to estimate. 

\subsubsection{Estimation of basis frequency and measurement residual phase}

The quadratic-basis ($\psi$) defined in Sec. II.C.1 is a function
of basis-frequency ($w_{0}$). This implies that finding the best-fit
coefficients to each PTS requires the a-priori knowledge of $w_{0}$\footnote{Note that $w_0$ is the RR which is the dominant frequency of RP.}.
Although this requirement seems to demand a crucial parameter a priori,
it will be shown in the subsequent sections that the inaccuracy in
the initial choice of $w_{0}$ does not affect the estimate of the
RP. Further, the exact state of the breathing of the subject at the
start of the video capture is unknown. This results in a residual
phase-lag between the estimated signal and the generating signal which
is to be compensated for. We compute a `proxy signal' that would simultaneously
estimate $w_{0}$ and compensate for the residual phase. Proxy signal
is a time-series whose value at a given time is obtained by taking
the element-wise dot product of intensities of all the pixels in the
video frame corresponding to that time, with respect to the intensities
of the pixels in the very first video frame. Mathematically, if $x(t)$
is the column vector obtained by stacking all the intensity values
in frame at time $t$, then the proxy signal $p(t)$ is defined as
$p(t)=(\mathbf{x}(1)\cdot\mathbf{x}(t))/(|\mathbf{x}(1)||\mathbf{x}(t)||)$.
$p(t)$ being the cosine of angle between two vectors, defines the
state of breathing in every frame with respect to the very first frame.
Also, it will be periodic and its fundamental frequency is used as
the basis frequency $w_{0}$. The residual phase correction is made
by projecting one period of the proxy signal on to the data elliptical
disk. Further the major axis of the data elliptical disk is reoriented
along the direction of the projected point corresponding to the proxy
signal. 

\subsubsection{Goodness-of-estimation (GoE) measure for parameter selection }

In section II.F, to handle the trade-off between estimating the higher
harmonics better and reducing the impact of noise, a GoE measure is
described to select the parameter deciding the part of the disk to
be included in $X^{+}$. Since the generating signal to be estimated
is periodic a measure that quantifies the closeness of a signal to
this behaviour will serve as a GoE measure. It is known that the Fourier
magnitude spectrum of a periodic signal is sparse and $l_{0}$ norm
(number of non-zero elements in a set) of the magnitude spectrum of
periodic signals should be lower than that for non-periodic signals.
Hence, for aggregation of points to be chosen in $X^{+}$, we start
from the periphery and choose the parameter (radius of exclusion,
See II.F.) that would result in the least $l_{0}$ norm. 

\section{Experiments and results}

The theory and the method proposed in Sec.II is validated using simulated
periodic signals and real-life breathing video data. The simulated
data serves the purpose of directly verifying the theoretical claims
by having a control over all the variables involved whereas the real-life
data is to demonstrate the usability of the method in real-life scenarios. 

\subsection{Simulated Data}

This data comprises of seven different generating signals of the form
$g(t)=\ensuremath{\sum_{k=1}^{N}G_{k}sin(w_{k}t+\theta)}$ as described
in Table I. 
\begin{table}[tbh]
\begin{centering}
\caption{Properties of the signals used to to simulate different cases.}
\par\end{centering}
\centering{}%
\begin{tabular}{|c|c|c|}
\hline 
Type of the signal $\left(g(t)\right)$ & $G_k$  & $f_k (deci\,\, Hz)$\tabularnewline
\hline 
\hline 
Single frequency  & ${1}$ & ${50}$\tabularnewline
\hline 
Two frequencies & ${1, \frac{1}{3}}$ & ${25, 75}$\tabularnewline
\hline 
Three frequencies & ${1, \frac{1}{3}, \frac{1}{2}}$ & ${20, 60, 120}$\tabularnewline
\hline 
Four frequencies & ${1, \frac{1}{6}, \frac{1}{8}, \frac{1}{12}}$ & ${25, 50, 100, 125}$\tabularnewline
\hline 
Sawtooth wave ($N=10$) & $\frac{1}{k}$ & $20*k$\tabularnewline
\hline 
Square wave ($N=10$) & $\frac{1}{2k - 1}$ & $20*(2k - 1)$\tabularnewline
\hline 
Triangle wave ($N=10$) & $\frac{(-1)^{k-1}}{(2k - 1)^2}$ & $20*(2k - 1)$\tabularnewline
\hline 
\end{tabular}
\end{table}
 Each $g(t)$ is passed through a set (5000 samples, $i$) of random
LTI systems connected in parallel to obtain $f_{i}(t)$ (Sec. II.E)
with $F_{i}(w_{k})=\mathbb{U}\sim[0,1]$ and $\phi_{i}=\mathbb{U}\sim[-\pi,\pi]$.
These responses, $f_{i}(t)$, are added with a noise process $n(t)\sim\mathbb{N}(0,\sigma^{2})$\footnote{Since no distributional assumption is made on the noise, any distribution would suffice.}
to yield $x_{i}(t)$. A part of each $x_{i}(t)$ of length corresponding
to $w_{0}$ is projected on to the quadratic basis $(\psi$) described
in Lemma 4 to obtain the elliptical disks of coefficients.

\subsubsection{Experiments and validation metrics }

We report three experiments as follows - (1) different amounts of
noise are added to each $g(t)$ that is estimated using the method
described and the normalized cross-correlation between $g(t)$ and
the estimated signals are studied, (2) The extent of validity of $l_{0}$norm
as the GoE measure is studied against the normalized cross-correlation
measured between $g(t)$ and the estimated signals and (3) sensitivity
of the method to the choice of the basis frequency ($w_{0}$) is studied
by comparing the correlation and the fundamental frequency of the
estimated signals (with an improper choice of $w_{0}$) with $g(t)$. 

\subsubsection{Results and discussion }

\begin{figure}[tbh]
\begin{centering}
\includegraphics[width=3.6in]{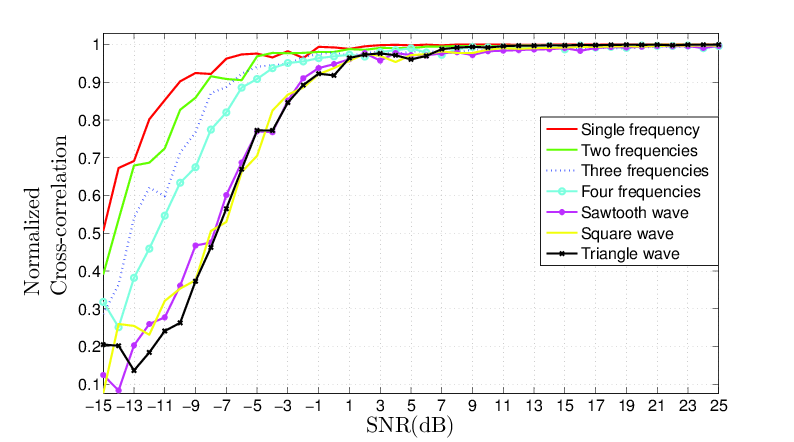}
\par\end{centering}
\caption{Correlation between the estimated signals and different $g(t)$ with
SNR ranging between -15 to 25 dB.\label{fig:exp1}}
\end{figure}

Fig. \ref{fig:exp1} depicts the cross-correlation between the estimated
signals and different $g(t)$ with SNR ranging between -15 to 25 dB.
 The threshold for radius of exclusion was determined by the $l_{0}$
norm GoE. It is seen that for all the signals, as SNR increases the
cross-correlation generally increases and saturates around 1 dB. However,
at SNRs lower than -2 dB, a signal with lower number of harmonics
achieves a certain cross-correlation before a signal that has higher
number of harmonics. It is seen that for all signals the cross-correlation
reaches 0.9 around -2 dB implying that this method can recover the
signal to a fairly good extent even when noise power is more than
that of the signal.

\begin{figure}[tbh]
\begin{centering}
\includegraphics[width=3.6in,height=1.5in]{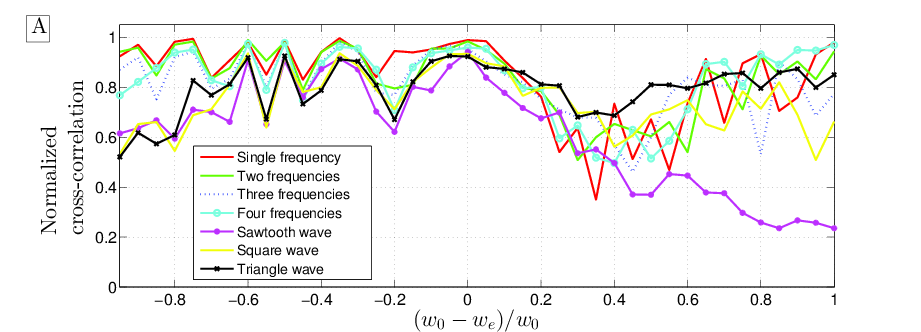}
\par\end{centering}
\begin{centering}
\includegraphics[width=3.6in]{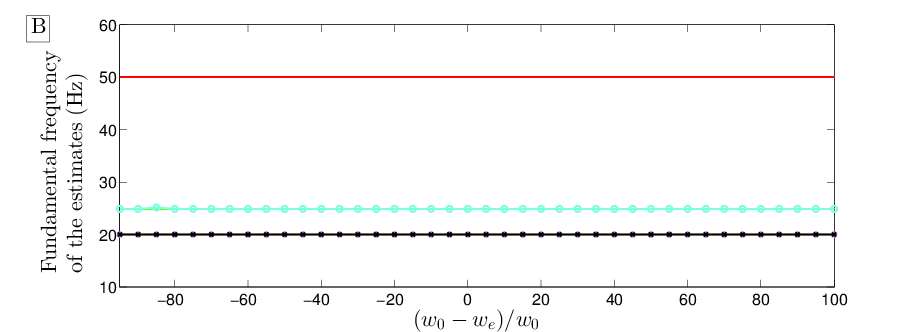}
\par\end{centering}
\caption{Analysis of sensitivity of the method to the choice of basis frequency.
(A): Normalized cross-correlation between the estimated signal and
$g(t)$ (B) Fundamental frequency of the estimated signal. Both (A)
and (B) are plotted as a function of of $\frac{(w_{0}-w_{e})}{w_{0}}$.
\label{fig:exp2}}
\end{figure}

In the next experiment, we fix the SNR at 0 dB and study the properties
of the estimated signal by varying the choice of basis-frequency ($w_{e}$)
between $0.05w_{0}$ and $2w_{0}$ where $w_{0}$ is the actual fundamental
frequency of a given $g(t)$. Fig. \ref{fig:exp2} (A) depicts the
normalized-cross correlation between the estimated signal and $g(t)$
as a function of $\frac{(w_{0}-w_{e})}{w_{0}}$. It is seen that good
estimates are obtained only around $w_{e}=w_{0}$ and estimates degrade
on either sides. This implies that the method is very sensitive to
the choice of $w_{0}$. However, the method can be easily tweaked
to circumvent this problem as evident from the following discussion:
Fig. \ref{fig:exp2} (B) depicts the fundamental frequency of the
estimated signals as a function of the same $\frac{(w_{0}-w_{e})}{w_{0}}$
as in Fig. 6 (A). It can be seen that the fundamental frequency of
all the estimated signals (taken to be the frequency at which the
magnitude Fourier spectrum peaks) are exactly the same as that of
the corresponding $g(t)$ (as inferred from Table I). This implies
that the peak of the magnitude spectrum of the estimated signals is
totally insensitive to the choice of basis-frequency and the lower
cross-correlation is due to the aggregation of wrong phase lags $(\phi_{i}$)
of the LTI channels that are selected in $X^{+}$. This is also supported
from the theory because to get a good estimate of the magnitude response
it is enough to satisfy criteria 1 and 2 listed in section II.F despite
violating criteria 3 and 4. A wrong choice of $w_{0}$ still leads
to an elliptical disk but with an improper orientation of $\phi_{i}$
with respect to the actual phase of $g(t)$, $\theta$. In this case
the proposed method still picks up the points required for an accurate
estimation of the magnitude response albeit distorting the shape of
the estimated signal due to the selection of LTI channels with improper
phase lags. This suggests that a simple way to circumvent the sensitivity
of the method to the choice of $w_{0}$ is to adopt a two-step procedure
where the initial step is to derive the actual $w_{0}$ (with any
initial choice of $w_{0}$) and in the next step is to use the proper
$w_{0}$ to estimate the morphology of $g(t)$. 

\begin{figure}[tbh]
\begin{centering}
\includegraphics[width=3.7in]{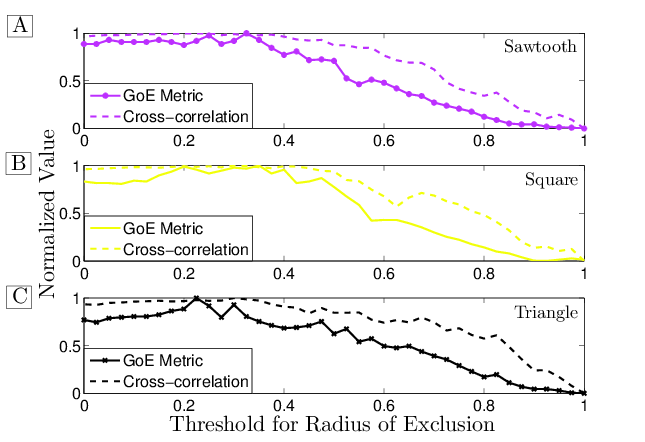}
\par\end{centering}
\caption{Validation of $l_{0}$ norm as the GoE metric for the selection of
optimal choice of the threshold used for selecting the radius of exclusion.
Values of inverse of $l_{0}$norm (GoE) of the estimated signals and
correlation between the estimated signal and $g(t)$ for three signals
- Sawtooth wave (A) and Square wave (B) and Triangle wave (C) at $-5$dB
SNR as a function of threshold for radius of exclusion.\label{fig:goe}}
\end{figure}

 In the proposed method, the optimal choice of the threshold used
for selecting the radius of exclusion (discussed in Sec. II.F) is
decided based on the GoE metric. In the last experiment, we validate
the proposed metric ($l_{0}$ norm of the magnitude spectrum of the
estimated signal) by comparing it against the cross-correlation measure.
Fig. \ref{fig:goe} depicts the values of inverse of $l_{0}$ norm
(GoE) of the estimated signals and cross-correlation between the estimated
signal and $g(t)$ for three signals: sawtooth, square and triangle
wave at $-5$dB SNR as a function of threshold for radius of exclusion.
It is to be noted that the value of threshold corresponding to unity
represents the selection of all points in one half of the ellipse.
The proposed method selects that threshold corresponding to the highest
GoE which estimates a signal with a very high correlation with $g(t)$
as seen from Fig.\ref{fig:goe}. This indicates that the defined measure
for GoE can be used as a proxy to determine the threshold in the practical
cases where the correlation measure cannot be computed due to the
unavailability of $g(t)$. 

\subsection{Real-life Dataset }

\begin{figure*}[tbh]
\begin{centering}
\includegraphics[width=6in]{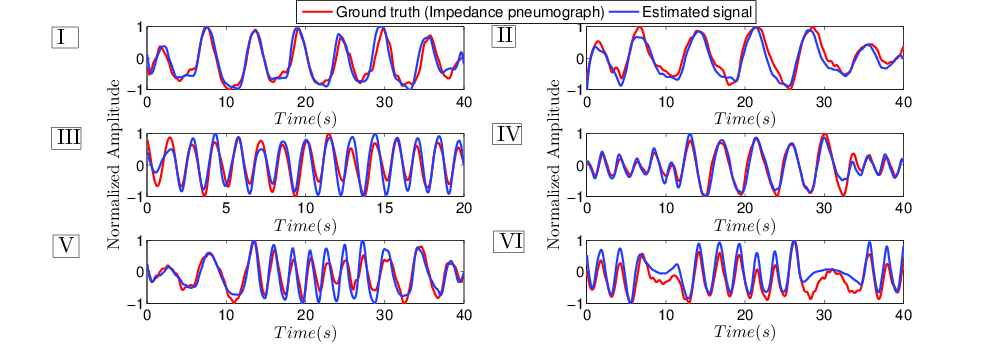}
\par\end{centering}
\caption{Depiction of six different validation experiments performed : Red
curve corresponds to normalized amplitudes of the impedance pneumogram
device attached to the subject and blue curves are the estimated signal.\label{fig: groud_truth_depic} }
\end{figure*}

The real-life dataset comprises respiration videos acquired from 31
healthy human subjects (for which institutional approval and subject-consent
were obtained) (10 female and 21 male) between ages of 21 - 37 (mean:
28). Six controlled breathing experiments (Fig. \ref{fig: groud_truth_depic})
(I) normal breathing, (II) deep breathing, (III) fast breathing, (IV)
normal-deep-normal breathing (sudden change in breathing volume),
(V) normal-fast-normal breathing (sudden change in breathing frequency),
(VI) episodes of breath hold, ranging between 13 - 150 (mean: 45)
seconds were performed by each subject. The subjects wore a wide variety
of clothing with different textures or no upper body clothing (two
subjects). Videos were simultaneously recorded from two cameras with
resolution of 640X480 pixels at a speed of 30 frames per second, one
from the ventral view (VGA) and other from lateral view (2MP) of the
subject under normal indoor illumination, each placed at a distance
of 3 ft from the subject. The subjects were asked to sit and breath
in patterns described above resulting in a total of ~2.5 hours of
recordings with approximately 2000 respiratory cycles for each side.
For validation, an impedance pneumograph (IP) device \cite{geddes1962impedance}
was connected through electrodes on the chest of the subject, which
estimates the RP and RR by quantifying the changes in electrical conductivity
of the chest due to respiratory air-flow. This device is routinely
used in patient monitors and other applications in which it is considered
a medical gold standard \cite{pacela1966impedance}. 

\subsubsection{Experiments and validation metrics }

\begin{figure}[tbh]
\begin{centering}
\includegraphics[width=3in]{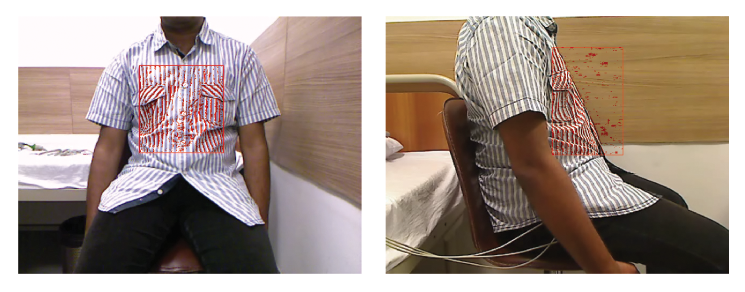}
\par\end{centering}
\caption{Depcition of an actual data scene with the selected memebership set
$X^{+}$ marked in red. \label{fig:Depcition-of-an}}

\end{figure}

Given a subject video, the proposed algorithm is applied to estimate
the RP and RR retrospectively. A typical frame (as shown in Fig. \ref{fig:Depcition-of-an}
) also consists of regions like background wall, that contain pixels
that are unaffected by respiration. An image gradient operation is
applied over a large rectangular window on two arbitrarily selected
frames at the beginning of the video that are spaced apart by the
minimum possible RR. Subsequent frames are pruned to contain only
those pixels with very high values of the gradient. This selects a
smaller region of the frame typically comprising the chest-abdomen
region of the subject. Note this operation is done only once on a
pair of frames at the beginning of the video. This is to reduce the
unnecessary processing of static pixels even though the proposed algorithm
does not demand the same. The initial estimate of basis-frequency
and the residual measurement phase are obtained using the proxy signal.
Quadratic coefficients are obtained for every pixel time series to
form the elliptical disk from which the optimum membership set $X^{+}$
and thus the RP are estimated. Once $X^{+}$ is assembled by tagging
PTS (which only involves computation of inner products), the algorithm
can be executed in real-time since the estimator only computes a pixel
average over $X^{+}$. Once the RP is obtained, RR is estimated from
the peak in the Fourier magnitude spectrum of the RP taken over a
window (typically between 10 and 15 seconds). 

The goal of this study is to estimate the morphology of the respiration
airflow signal (RP) and not the actual airflow. Further, the IP device
also does not directly provide the volumetric information albeit it
has been shown to provide the actual airflow information with proper
calibration \cite{geddes1962impedance}. Thus we use the Pearson correlation
coefficient \cite{benesty2009pearson} between the normalized signal
obtained from the IP representing the ground truth, GT, and the normalized
estimated RP as a measure quantifying the closeness of two signals.
This measure lying between -1 and 1 quantifies the closeness of two
temporal signals with unity referring to the maximum agreement. RR
measurements are validated through the linear regression between GT
and the estimated RR values. Further since RR is a frequency measurement
that can be exactly obtained from both the signals, the exact agreement
is quantified using the Bland-Altman plots \cite{bland1986statistical}. 

\subsubsection{Results and discussion }

\begin{figure*}[!t]
\begin{centering}
\includegraphics[width=6.2in]{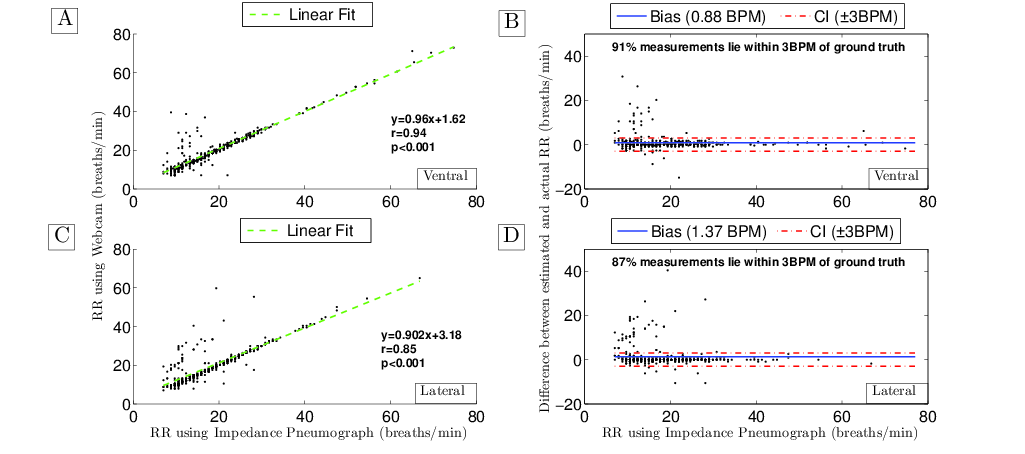}
\par\end{centering}
\caption{Correlation and agreement of the estimated signals with $g(t)$: (A)
and (C) show the degree of linear relationship between RR measurements
using webcam and IP device. (B) and (D) are the Bland-Altman plots
showing the difference in the RR values with a confidence interval
of $\pm3$ BPM (CI) obtained from the two methods against the ground
truth. For the ventral video acquisition, in (A) we observe that the
Pearson correlation coefficient ($r$ ) is 0.94 with $p<0.001$, which
shows a strong positive correlation between the measurements. \label{fig:Correlation-and-agreement}}
\end{figure*}

\begin{figure}[tbh]
\begin{centering}
\includegraphics[width=3.7in]{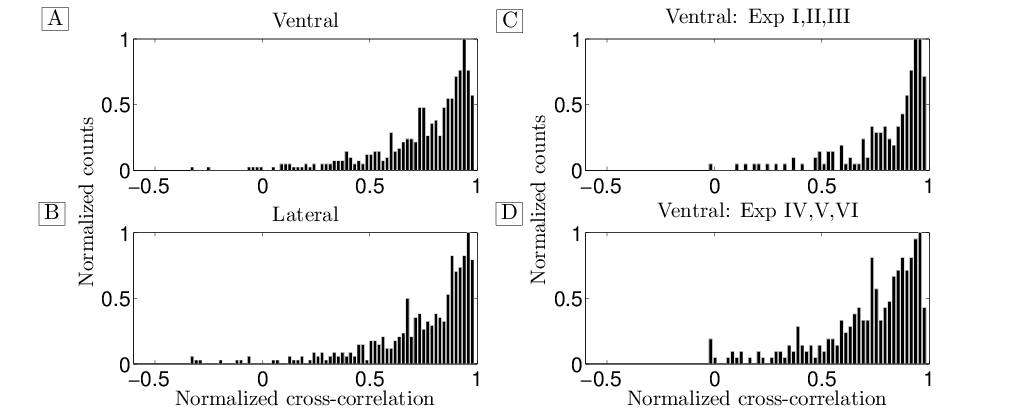}
\par\end{centering}
\caption{Histograms of signal correlation between the estimated RP and the
GT. (A) RP ventral Vs. GT, (B) RP lateral Vs. GT, (C) RP ventral Vs.
GT for experiments (I, II, III), (D) RP ventral Vs. GT for experiments
(IV, V VI).\label{fig:Histograms-of-signal}}
\end{figure}

Figure \ref{fig:Correlation-and-agreement} depicts the correlation
and agreement of the estimated signals with $g(t)$: (A) and (C) show
the degree of linear relationship between RR measurements using webcam
and IP device. For the ventral video acquisition, in Fig. \ref{fig:Correlation-and-agreement}
(A), it is observed that the correlation coefficient ($r$ ) is 0.94
with $p<0.001$, which shows a strong positive correlation between
the measurements. Also, Bland-Altman plots in Fig. \ref{fig:Correlation-and-agreement}(B)
shows that the RR measurement through webcam has an acceptable average
agreement (very low bias of 0.88) with the ground truth with $91\%$
of the measurements within 3BPM of the ground truth. The median of
deviation between the estimated values and the GT values is zero and
the measurements that are outside of the confidence interval (CI,
defined as $\pm$3 BPM of the ground truth values) are often higher
than the actual RR. These are the cases where there are high-frequency
repetitive and densely patterned textures. For the case of lateral
video acquisition, in Fig. \ref{fig:Correlation-and-agreement} (C)
and (D), a correlation coefficient ($r$ ) of 0.85 is observed with
$p<0.001$ and $87\%$ of the measurements lie within 3 BPM of the
ground truth. These numbers are lesser than those for the frontal
view because the lateral view typically has much fewer members in
$X^{+}$.
 Figure \ref{fig:Histograms-of-signal} depicts the histograms of
the signal correlation measure between the estimated RP and GT for
different cases. It is seen that the mode of the histogram for all
cases is around 0.9 with a negative skew indicating that majority
of the estimated signals agree well with the GT. 
Also, it is seen that the skewness of the histogram for experiments
IV, V and VI is worse than that for experiments I, II and III as indicated
in Fig. \ref{fig:Histograms-of-signal} (C and D). This is because
the generating signals corresponding to experiments IV, V and VI have
time-varying frequency components. However, note that for the case
in which only spectral magnitudes vary with time but not frequency
values, the theoretical results and performance of the proposed method
remains unaltered. 

\textcolor{black}{Figure \ref{fig:Depcition-of-an} depicts a scene
from one of the experiments on a subject, with the selected membership
set ($X^{+}$) marked with red dots. It is seen that the selected
pixels are not geographically contiguous and mostly in the abdominal-thoracic
region of the subject where the respiratory movement is significantly
manifested. It is also possible that there can be a small number of
very low SNR pixels that get assigned to $X^{+}$ because of the effect
of noise-cloud explained in Sec. II. E. The LTI channels corresponding
to these pixels do not contain significant respiratory information
(for instance, red points on the wall in lateral view of Fig. \ref{fig:Depcition-of-an}).
However, these points often do not alter the estimate as, often, they
are very small in number compared to the pixels that have significant
signal components and thus get nullified while ensemble averaging.
The algorithm allows further control over such instances through the
choice of parameter GoE and hence radius of exclusion ($r_{e}$). }

The aforementioned performance of the algorithm seems significant
given that the experiments involve the following: (i) random textured
clothing on subjects, (ii) camera of two different resolutions and
positions, (iii) six different breathing patterns. In conclusion,
it is observed that the proposed algorithm offers a good estimate
of the RP (and RR) if the camera is placed in the ventral position
with a clothing that has a texture with some region of similar patterns. 

\subsection{\textcolor{black}{Robustness aspects }}

\textcolor{black}{In this section, we discuss the robustness of the
proposed method for the cases where there are deviations from the
assumed models and discuss its noise tolerance for the application
considered. }
\begin{enumerate}
\item \textcolor{black}{Amplitude modulation - Suppose the generating signal
$g(t)$ is modulated with a slowly time-varying modulating signal
$m(t)$, that is, $g(t)=\ensuremath{m(t)\sum\limits _{k=1}^{N}G_{k}sin(w_{k}t+\theta)}$.
For such signals, it can be easily shown that the results developed
in Sec. II remain valid. An intuitive understanding of this fact may
be inferred from noticing that the estimate proposed here has an arbitrariness
in the amplitude scale of the estimated signal. Thus, in case of amplitude
modulation, the estimated signal remains to be the actual generating
signal up to an arbitrary constant scaling factor. This fact is corroborated
with the results of experiments IV, V and VI on the real-life data
set, which have amplitude modulating components in them. }
\item \textcolor{black}{Frequency modulation - Suppose the generating signal
is frequency modulated, that is, $g(t)=\sum\limits _{k=1}^{N}G_{k}sin(w_{k}(t)t+\theta)$
where $w_{k}(t)$ is slowly varying. For this case, the results developed
in Sec. II, do not extend directly. This is because of two reasons:
(a) frequency modulated signals do not adhere to the periodicity assumption
that is imposed on the generating signal during the development of
the theory, (b) the basis vectors on which the signals are projected
are a function of fundamental frequency required to be a constant
(Eq. \ref{eq:astar} and \ref{eq:bstar}). However the proposed method
can be modified to retain approximate validity for the case where
the generating signals are quasi-periodic, that is, $w_{k}(t)$ is
slowly varying.\footnote{ {This is a reasonable assumption in the case of many real-world bio signals which possess constant base-frequency for short durations of time.}}A
straightforward extension of the proposed method for this category
of signals is to recompute $X^{+}$ at regular short intervals (governed
by the assumed interval of the quasi-periodicity). This method is
shown to yield considerably good estimates of the signal morphology
in the cases of the real-data for the cases of experiments IV, V and
VI where the generating signal resembles a quasi-stationary frequency
modulated signal. }
\item \textcolor{black}{Tolerance to sporadic body-movement and background
disturbances - In practical scenarios, there will be movements in
the human body that do not correspond to respiratory motion. In addition,
there can also be movements caused by objects other than the human
body. If such movements impact only the pixels that are not contained
in $X^{+}$, the performance of the method remains unaltered. In the
event such movement influences pixels within $X^{+}$, it has been
observed that the ensemble averaging retains robust performance provided
such motion impacts a smaller fraction of $X^{+}$. This fact has
also been corroborated with the results on the real-life data set
where the subjects were not strictly constrained to be still but there
were asked to breath sitting in a relaxed posture. }
\end{enumerate}
\textcolor{black}{Given the aforementioned discussions, some of the
major merits of the proposed method may be listed as follows - (a)
it is a generic framework for blind-deconvolution of SIMO systems
driven by periodic inputs without the need for estimating the underlying
channel responses, noise characteristics or error minimization making
the method asymptotically exact when there is no noise, (b) when applied
to respiration pattern estimation, this method selects a set of most-relevant
pixels that would estimate the signal which need not be geographically
continuous and thus does not critically depend on ROI selection, camera
orientation and texture of the surface, (c) it is computationally
inexpensive and thus can be implemented in real-time. Nevertheless,
the proposed estimator is limited by (a) its inability to quantify
true magnitude of the generating signal and (b) its non-applicability
to the class of signals with rapidly varying time-frequency components. }

\section{Conclusion and future work}

In this paper, we proposed a generic blind deconvolution framework
to extract periodic signals from videos. A video is modeled as an
ensemble of LTI measurement channels all driven by a single generating
signal. No assumptions are made on the characteristics of the individual
channels except for IID randomness. A simple ensemble averaging over
a carefully selected membership set is proposed as an effective estimator
which is shown to converge to the generating signal under minimally
restrictive assumptions. A method for grouping the channels to obtain
the optimal membership set based on the location of the coefficients
of the quadratic fits of the LTI channel responses is described. This
framework is applied on the problem of non-contact respiration pattern
estimation using videos and it is shown to yield comparable results
with a medical gold-standard device namely impedance pneumograph.
Our future work is aimed at extending this framework to (i) deal with
signals having rapidly varying time-frequency components, (ii) estimate
other relevant biomedical signals from video and (iii) deal with significant
sources of motion other than the one caused by the desired source. 

\appendices{}

\section*{Acknowledgment}

We acknowledge the support provided by our colleagues Dr. Satish P
Rath, Tejas Bengali and Himanshu J Madhu pertaining to various aspects
of the work.

\bibliographystyle{IEEEtran}
% Generated by IEEEtran.bst, version: 1.14 (2015/08/26)

\end{document}